\documentclass{article}

\usepackage[preprint,nonatbib]{nips_2018}

\usepackage{microtype}
\usepackage{amsmath}
\usepackage{amsfonts}
\usepackage{graphicx}
\usepackage{subfigure}
\usepackage{siunitx}
\usepackage{colortbl}
\usepackage{makecell}
\usepackage{wrapfig}
\usepackage{caption}
\usepackage{xcolor}
\usepackage{booktabs} % for professional tables

\newcommand*{\ArXiv}{}

\usepackage{hyperref}

\title{
  Discovery of Latent 3D Keypoints via\\
  End-to-end Geometric Reasoning
}

\usepackage{bbm}
\usepackage{amsthm}
\usepackage{amsmath}
\usepackage{amssymb}
\usepackage{color}
\usepackage{xspace}
\usepackage{pgfplots, pgfplotstable}
\usepackage{multirow}

\def\shortname{KeypointNet\xspace}

\newcommand{\diag}{\mathop{\mathrm{diag}}}
\newcommand{\svd}{\mathop{\mathrm{SVD}}}

\newcommand{\R}[1]{R}%^{#1}}
\newcommand{\G}[1]{G}%^{#1}}

\newcommand{\Ad}[1]{A}

\newcommand{\CC}[1]{C}

\def\eg{{\em e.g.,}}
\def\ie{{\em i.e.,}}

\def\etal{{\em et al.}\xspace}
\def\vs{{\em vs.}\xspace}

\newcommand{\figref}[1]{Figure~\ref{#1}}

\newcommand{\secref}[1]{Section~\ref{#1}}

\newenvironment{packed_item}{
\begin{itemize}
  \setlength{\itemsep}{2.5pt}
  \setlength{\parskip}{0pt}
  \setlength{\parsep}{0pt}
}{\end{itemize}}

\def\link{\href{https://keypointnet.github.io/}{\em keypointnet.github.io}\xspace}

\author{
\hspace{-0.7em}\begin{tabular}{c@{\hspace*{.35cm}}c@{\hspace*{.35cm}}c@{\hspace*{.35cm}}c}
  Supasorn Suwajanakorn\textsuperscript{$\triangleright$}\thanks{Work done while S. Suwajanakorn was a member of the Google AI Residency program (g.co/airesidency).} & Noah Snavely\textsuperscript{$\triangleleft$} & Jonathan Tompson\textsuperscript{$\triangleleft$} & Mohammad Norouzi\textsuperscript{$\triangleleft$}\\
  \end{tabular}\\[.1cm]
    \texttt{supasorn@vistec.ac.th, \{snavely,\:tompson,\:mnorouzi\}@google.com}\\
    \textsuperscript{$\triangleright$}Vidyasirimedhi Institute of Science and Technology\hspace*{0.45cm}
    \textsuperscript{$\triangleleft$}Google AI\\
}

\long\def\comment#1{}

\begin{document}

\maketitle

\vspace{-.5cm}
\begin{abstract}
\vspace{-.2cm}
This paper presents {\em \shortname}, an end-to-end geometric
reasoning framework to \emph{learn} an optimal set of {\em
category-specific} 3D keypoints, along with their detectors.  Given a
single image, \shortname extracts 3D keypoints that are optimized for
a downstream task.  We demonstrate this framework on 3D pose
estimation by proposing a differentiable objective that seeks the
optimal set of keypoints for recovering the relative pose between two
views of an object. Our model discovers geometrically and semantically
consistent keypoints across viewing angles and instances of an object
category. Importantly, we find that our end-to-end framework using no
ground-truth keypoint annotations outperforms a fully supervised
baseline using the same neural network architecture on the task of
pose estimation.  The discovered 3D keypoints on the car, chair, and
plane categories of ShapeNet~\cite{chang2015shapenet} are visualized
at \link.

\end{abstract}

\vspace{-.3cm}
\section{Introduction}
\vspace{-.1cm}
\label{introduction}
Convolutional neural networks have shown that jointly optimizing
feature extraction and classification pipelines can significantly improve
object recognition~\cite{lenet,alexnet}.  That being said, current
approaches to geometric vision problems, such as 3D
reconstruction~\cite{phototourism} and shape
alignment~\cite{li2009robust}, comprise a separate {\em keypoint}
detection module, followed by geometric reasoning as a post-process.
In this paper, we explore whether one can benefit from an {\em
end-to-end} geometric reasoning framework, in which keypoints are
jointly optimized as a set of {\em latent variables} for a downstream
task.
% is able to discover a set of {\em latent} keypoints based on 3D pose annotations (2)

% function leads to better performance, even though
% it is not making use of strong supervision at the keypoint detection stage.

% Past generations of object recognition pipelines comprised hand-crafted feature extraction (\eg~\cite{sift, hog}),
% followed by classification. Convolutional neural networks~\cite{lenet} jointly optimize the feature extraction and classification modules.
% This results in an impressive performance gains on object recognition~\cite{alexnet, resnet} and segmentation~\cite{fcn, he2017mask}.
% This paper examines whether a {\em sparse} set of keypoints or landmarks
% one can obtain similar benefits from an end-to-end optimization of representations based on a {\em sparse}
% set of keypoints or landmarks, often used in geometric reasoning applications, such as 3D reconstruction~\cite{phototourism} and 3D shape retrieval.

% From such data, one learns a supervised mapping from images to keypoint positions.
% followed by geometric reasoning (\eg~PnP algorithm~\cite{lepetit2008pnp}) on the sparse set of keypoint detections to recover the 3D pose or the camera angles from a given image.
% multiple views.

Consider the problem of determining the 3D pose of a car in an image.
A standard solution first detects a sparse set of category-specific
keypoints, and then uses such points within a geometric reasoning
framework (\eg~a PnP algorithm~\cite{lepetit2008pnp}) to recover the
3D pose or camera angle.  Towards this end, one can develop a set of
keypoint detectors by leveraging strong supervision in the form of
manual keypoint annotations in different images of an object category,
or by using expensive and error prone offline model-based fitting
methods.  Researchers have compiled large datasets of annotated
keypoints for faces~\cite{sagonas2016300}, hands~\cite{tompson14tog},
and human bodies~\cite{andriluka14cvpr, lin2014microsoft}.  However,
selection and consistent annotation of keypoints in images of an
object category is expensive and ill-defined.  To devise a reasonable
set of points, one should take into account the downstream task of
interest. Directly optimizing keypoints for a downstream geometric
task should naturally encourage desirable keypoint properties such as
distinctiveness, ease of detection, diversity, {\em etc.}
%Further, not all keypoints are equally easy to detect; the ease of
%detectability of keypoints across different views plays an important
%role in the final performance of a downstream task.

This paper presents {\em \shortname}, an end-to-end geometric
reasoning framework to learn an \emph{optimal} set of
category-specific 3D keypoints, along with their detectors, for a specific downstream task. Our novelty stands in contrast to prior work that learns latent keypoints through an arbitrary proxy self-supervision objective, such as reconstruction \cite{zhang2018unsupervised,hinton2011transforming}.
%This paper presents {\em \shortname}, a
%novel framework to \emph{discover} an optimal set of keypoints for a
%given object class, along with learned detectors to predict these
%keypoints from 2D images.
Our framework is applicable to any downstream task represented by an
objective function that is differentiable with respect to keypoint positions.
% This paper presents {\em \shortname}, a novel framework for learning category specific keypoints, along with their detectors. The framework applies to any downstream task with a differentiable objective function in terms of the keypoint positions.
We formulate 3D pose estimation as one such task, and our key
technical contributions include (1) a novel differentiable pose estimation objective and (2) a multi-view consistency loss function. The pose objective seeks optimal keypoints for recovering
the relative pose between two views of an object. The multi-view
consistency loss encourages consistent keypoint detections across 3D
transformations of an object. Notably, we propose to detect \emph{3D}
keypoints (2D points with depth) from individual 2D images and
formulate pose and consistency losses for such 3D keypoint detections.

% We also propose generic keypoint constraints including multi-view consistency and silhouette consistency.
% , learned from object masks, to ensure proper keypoint characteristics. 

%manifest the minimal requirements of a reasonable set of stable 3D keypoints.
%Unlike prior work, we do not require keypoint annotations, and naturally incorporate the ease of detectability of the keypoints while optimizing them.

%

% * table is the largest, and we use shapenet subclass. 
We show that \shortname discovers geometrically and semantically
consistent keypoints across viewing angles as well as across object
instances of a given class. Some of the discovered keypoints
correspond to interesting and semantically meaningful parts, such as
the wheels of a car, and we show how these 3D keypoints can infer
their depths without access to object geometry.  We conduct three sets
of experiments on different object categories from the ShapeNet
dataset~\cite{chang2015shapenet}. We evaluate our technique against a
strongly supervised baseline based on manually annotated keypoints on
the task of relative 3D pose estimation.
%To train this baseline, we first collect keypoint annotations by asking Mechanical Turk workers to locate reference keypoints based on Pascal3D+~\cite{xiang2014beyondpascal} in our training images, and use these keypoint locations as a direct supervision. 
%We ask Mechanical Turk workers to annotate
%a number of keypoints in different images of various objects based on reference annotations from Pascal3D+~\cite{xiang2014beyondpascal}.
%Using a graphics pipelines, we map the annotations onto the underlying 3D shapes and synthesize many annotated images of an object category
%to train a supervised 3D keypoint detector in a fully supervised way.
%We compare \shortname using no keypoint supervision with the fully supervised baseline on the task
%of relative 3D pose estimation from a pair of views. 
Surprisingly, we find that our end-to-end framework achieves
significantly better results, despite the lack of keypoint
annotations.
% We conclude that end-to-end optimization matters more than keypoint annotation for our experiments.
%We plan to apply \shortname to other downstream vision tasks in the future.

% Further, we advocate the use of 3D coordinates for keypoint positions,~\ie~not only we detect $x$ and $y$ coordinates for any keypoint, but also predict the corresponding depth.
% Confirming the recent developments on depth estimation from a single image, we find that detecting the depth is possible, particularly when the keypoints are detected jointly.
% One benefit of using 3D coordinates is that we can even detect occluded keypoints.
% Importantly, using 3D keypoints from a single view, one can reason about the 3D pose and the 3D shape of the object.

\vspace{-.2cm}
\section{Related Work}
\vspace{-.1cm}
\label{related}
%%%%%%%%%%%%%%%%%%%%%%%%%%%%%%%%%%%%%%%%%%%%%%%%%%%%%%%%%%%%%%%%%%%%%%%%%%%
% \subsection{Human Keypoint Detection}

Both 2D and 3D keypoint detection are long-standing problems in
computer vision, where keypoint inference is traditionally used as an
early stage in object localization
pipelines \cite{lepetit2006keypoint}. As an example, a successful
early application of modern convolutional neural networks (CNNs) was
on detecting 2D human joint positions from monocular RGB images. Due
to its compelling utility for HCI, motion capture, and security
applications, a large body of work has since developed in this joint
detection domain~\cite{toshev2014deeppose, tompson2014joint,
pishchulin16cvpr, newell2016stacked, yang2017learning,
papandreou2017towards, huang2017coarse, he2017mask}.

% Removed wei2016convolutional from list above.

More related to our work, a number of recent CNN-based techniques have
been developed for 3D human keypoint detection from monocular RGB
images, which use various architectures, supervised objectives, and
3D structural priors to directly infer a predefined set of 3D joint
locations~\cite{VNect_SIGGRAPH2017, mehta2017monocular,
chen2017adversarial, mehta2017single, guler2018densepose}. Other
techniques use inferred 2D keypoint detectors and learned 3D priors to
perform ``2D-to-3D-lifting''~\cite{ramakrishna2012reconstructing,
chen20173d, zhou2016sparseness, martinez2017simple} or find
data-to-model correspondences from depth
images \cite{pons2015metric}. 
Honari \etal~\cite{honari2018improving} improve landmark localization by incorporating semi-supervised tasks such as attribute prediction and equivariant landmark prediction.
In contrast, our set of keypoints is
not defined {\em a priori} and is instead a latent set that is
optimized end-to-end to improve inference for a geometric estimation
problem.  A body of work also exists for more generalized, albeit
supervised, keypoint detection, \eg~\cite{NIPS2012_4680,
wu2016single}.

Enforcing latent structure in CNN feature representations has been
explored for a number of domains. For instance, the {\em capsule}
framework~\cite{hinton2011transforming} and its
variants~\cite{sabour2017dynamic,hinton2018matrix} encode activation
properties in the magnitude and direction of hidden-state vectors and
then combine them to build higher-level features.
%, by incorporating a novel routing-by-agreement algorithm to determine the connection of lower-level capsules to those higher in the network. 
The output of our \shortname can be seen as a similar form of latent
3D feature, which is encouraged to represent a set of 3D keypoint
positions due to the carefully constructed consistency and relative
pose objective functions.

Recent work has demonstrated 2D correspondence matching across
intra-class instances with large shape and appearance variation. For
instance, Choy \etal~\cite{choy2016universal}
%proposes a fully-convolutional architecture and 
use a novel contrastive loss based on appearance to encode geometry and
semantic similarity. Han \etal~\cite{han2017scnet} propose a novel
SCNet architecture for learning a geometrically plausible model for 2D
semantic correspondence.
Wang \etal~\cite{wang2017multi} rely on deep features and perform a multi-image matching across an image collection by solving a feature selection and labeling problem.
%In this work, region proposals are used for initial match primitives and geometric consistency is incorporated directly in the SCNet loss function.
Thewlis \etal~\cite{thewlis2017unsupervised} use ground-truth
transforms (optical flow between image pairs) and point-wise matching
to learn a dense object-centric coordinate frame with viewpoint and
image deformation invariance. Similarly,
Agrawal \etal~\cite{agrawal2015learning} use egomotion
% and camera transformation
prediction between image pairs to learn semi-supervised feature
representations, and show that these features are competitive with
supervised features for a variety of tasks. 

Other work has sought to learn latent 2D or 3D features with varying
amounts of supervision. Arie-Nachimson \&
Basri \cite{constructing_implicit_iccv_09} build 3D models of rigid
objects and exploit these models to estimate 3D pose from a 2D image
as well as a collection of 3D latent features and visibility
properties. Inspired by cycle consistency for learning
correspondence \cite{huang2013consistent,zhou2015multi},
Zhou \etal~\cite{zhou2016learning} train a CNN to predict
correspondence between different objects of the same semantic class by
utilizing CAD models. Independent from our work,
Zhang \etal~\cite{zhang2018unsupervised} discover sparse 2D landmarks
of images of a known object class as explicit structure representation
through a reconstruction objective.  Similarly, Jakab and Gupta  \etal~\cite{jakab2018conditional} use conditional image generation and reconstruction objective to learn 2D keypoints that capture geometric changes in training image pairs. 
Rhodin \etal~\cite{rhodin2018unsupervised} uses a multi-view consistency loss, similar to ours, to infer 3D latent variables specifically for human pose estimation task.
In contrast to \cite{zhou2016learning, zhang2018unsupervised, jakab2018conditional, rhodin2018unsupervised}, our latent keypoints are optimized
for a downstream task, which encourages more directed
keypoint selection. By representing keypoints in true physical 3D structures, our method can
even find occluded correspondences between images with large
pose differences, \eg~large out-of-plane rotations.

Approaches for finding 3D correspondence have been investigated.
Salti \etal~\cite{salti2015learning} cast 3D keypoint detection as a binary
classification between points whose ground-truth similarity label is
determined by a predefined 3D descriptor.
%This general descriptor allows their architecture to find correspondences across object classes with varying topology. 
Zhou \etal~\cite{zhou2017unsupervised} use view-consistency as a supervisory
signal to predict 3D keypoints, although only on depth
maps. Similarly, Su \etal~\cite{su2015render} leverage synthetically rendered
models to estimate object viewpoint by matching them to real-world
image via CNN viewpoint embedding. Besides keypoints, self-supervision based on geometric and motion reasoning has been used to predict other forms of output, such as 3D shape represented as blendshape coefficients for human motion capture \cite{tung2017self}.

\vspace{-.2cm}
\section{End-to-end Optimization of 3D Keypoints} \label{overview}
\vspace{-.1cm}
\begin{figure} \center
%\begin{tabular}{@{}cc@{}}
%Training&\\
%\multirow{3}{*}{\hspace*{-.2cm}\includegraphics[width=0.63\textwidth]{figures/keypoints_train.pdf}} & \\
%& Inference\\
%&\hspace*{-.2cm}
%{
\includegraphics[width=0.95\textwidth]{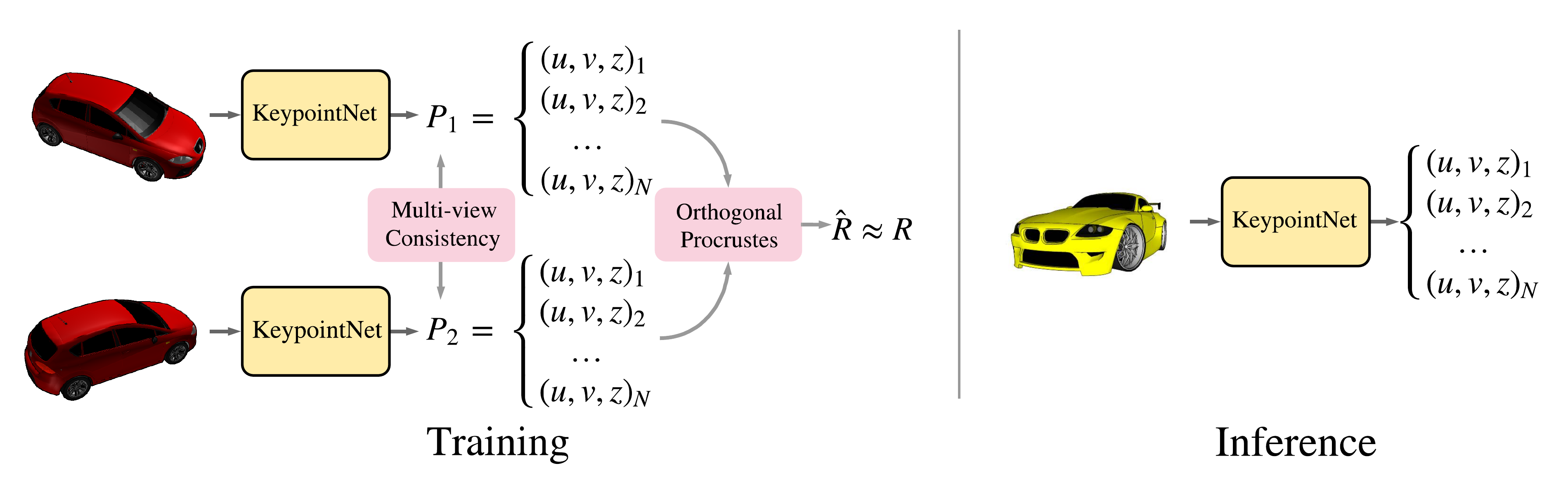}
%}\vspace*{1.2cm}\\[-.1cm]
%\end{tabular}
\vskip -.6em
\caption{During training, two views of the same object are given as
  input to the \shortname. The known rigid transformation $(R,t)$
  between the two views is provided as a supervisory signal.  We
  optimize an ordered list of 3D keypoints that are consistent in both views and enable recovery of the transformation.  During inference,
  \shortname extracts 3D keypoints from an individual input image.
\label{fig:overview}}
\vspace*{-.4cm}
\end{figure}

\comment{
\begin{figure} \center
\includegraphics[width=0.95\textwidth]{figures/pipeline.pdf}
\vskip -.6em
\caption{During training, two views of the same object are given as
  input to the \shortname. The known rigid transformation $(R,t)$
  between the two views is provided as a supervisory signal.  We
  optimize an ordered list of 3D keypoints that are consistent in both
  views and enable recovery of the transformation.  During test,
  \shortname extracts 3D keypoints from individual input images.
\label{fig:overview}}
\end{figure}
}

Given a single image of a known object category, our model predicts an
ordered list of 3D keypoints, defined as pixel coordinates and
associated depth values. Such keypoints are required to be
geometrically and semantically consistent across different viewing
angles and instances of an object category
(\eg~see~\figref{fig:cpcs}). Our \shortname has $N$ heads that extract
$N$ keypoints, and the same head tends to extract 3D points with the
same semantic interpretation. These keypoints will serve as a building
block for feature representations based on a sparse set of points,
useful for geometric reasoning and pose-aware or pose-invariant object
recognition~(\eg~\cite{sabour2017dynamic}).

%% Our
%% key insight is that one can optimize the keypoint detectors end-to-end
%% for a downstream task.

In contrast to approaches that learn a supervised mapping from images
to a list of annotated keypoint positions, we do not define the
keypoint positions \emph{a priori}. Instead, we jointly optimize
keypoints with respect to a downstream task.  We focus on the task of
{\em relative pose estimation} at training time, where given two views
of the same object with a known rigid transformation $T$, we aim to
predict optimal lists of 3D keypoints, $P_1$ and $P_2$ in the two views
that best match one view to the other (\figref{fig:overview}). We
formulate an objective function $O(P_1, P_2)$, based on which one can
optimize a parametric mapping from an image to a list of keypoints. Our
objective consists of two primary components:
%%%%%%%%%%%%%%%%%%%%%%%%%%
%to estimate the relative rigid transformation that relates one view to the other (\figref{fig:overview}).
%During training, we are given two views of the same object with a known rigid transformation $T$. 
%Given two sets of $N$ 3D keypoint positions detected in the two views (denoted $P$ and $P'$),
% , denoted
% $P = \{p_i \equiv (u_i, v_i, z_i)\}_{i=1}^N$
% and
% $P' = \{p_i' \equiv (u_i', v_i', z_i')\}_{i=1}^N$, 
%%%%%%%%%%%%%%%%%%%%%%%%%%
% ** MOVED to review in braintex
%\mohammad{int this section let's only talk about Procrustes loss and other possible alternatives.
%Let's move multi-view consistency to the next section where we talk about keypointness characteristics}
%%%%%%%%%%%%%%%%%%%%%%%%%%
\begin{packed_item}
\item A {\em multi-view consistency} loss that measures the discrepancy between
      the two sets of points under the ground truth transformation.
\item A {\em relative pose estimation} loss,
      which penalizes the angular difference between
      the ground truth rotation $R$ \vs
      the rotation $\hat{R}$ recovered from
      $P_1$ and $P_2$ using orthogonal procrustes.
\end{packed_item}
%%%%%%%%%%%%%%%%%%%%%%%%%%
We demonstrate that these two terms allow the model to discover
important keypoints, some of which correspond to semantically
meaningful locations that humans would naturally select for different
object classes. Note that we do not directly optimize for keypoints
that are semantically meaningful, as those may be sub-optimal for
downstream tasks or simply hard to detect. In what follows, we first
explain our objective function and then describe the neural
architecture of \shortname.

{\bf Notation.} Each training tuple comprises a pair of images $(I, I')$ of the same object from different viewpoints, along with their relative rigid transformation 
% $T \in \mathbb{SE}(3)$ 
$T \in SE(3)$, which transforms the underlying 3D shape from $I$ to $I'$. $T$ has the following matrix form:
\begin{align}\label{eq:t}
    T = \begin{bmatrix}
    {R}^{3 \times 3}  & t^{3 \times 1} \\
    {0} & 1
\end{bmatrix}~,
\end{align}
where ${R}$ and $t$ represent a 3D rotation and translation
respectively.  We learn a function $f_\theta(I)$, parametrized by
$\theta$, that maps a 2D image $I$ to a list of 3D points $P = (p_1,
\ldots, p_N)$ where $p_i \equiv (u_i, v_i, z_i)$, by optimizing an
objective function of the form $O(f_\theta(I), f_\theta(I'))$.

\comment{
Optionally, we have access to the orientation of the object in each image represented as a binary flag indicating left- or right-facing to help deal with visual ambiguities due to object symmetry. The total loss is a weighted sum of 4[or 6?] differentiable losses:
\begin{align}
    L &= \alpha_{con}L_{con} + \alpha_{pro}L_{pro} + \alpha_{key}L_{key}+ [\alpha_{sym}L_{sym}]
\end{align}
which 
} % end \comment

\subsection{Multi-view consistency}

%For an individual 3D keypoint to track the same object part,
%we need to ensure that the keypoints are consistent across different views, \ie that the predicted 2D %locations and depths are geometrically consistent with the known relative pose.
%%%% 
\comment{
We thus propose one kind of ``metric'' losses for keypoint selection, called Procrustes loss, which seeks the best set of keypoints for recovering the relative pose between the given training pair.
This loss naturally encourages keypoint separation, as having all keypoints in the same location would be degenerate in determining the rotation.
} % end \comment
%%%%
% The primary goal of the consistency loss is to ensure that a set of keypoints predicted in the first image
% is consistent with a second set of keypoint predictions.
%This is the goal of our multi-view consistency loss.

The goal of our multi-view consistency loss is to ensure that the keypoints track consistent parts across different views. Specifically, a 3D keypoint in one image should project onto the same pixel location as the corresponding keypoint in the second image.
For this task, we assume a perspective camera model with a known global focal length $f$. Below, we use $[x, y, z]$ to denote 3D coordinates, and $[u, v]$ to denote pixel coordinates.
The projection of a keypoint $[u, v, z]$ from image $I$ into image $I'$ (and vice versa) is given by the projection operators:
\begin{eqnarray*}%\label{eq:project} EDIT(tompson): label added by mistake?
    [\hat{u}, \hat{v}, \hat{z}, 1]^{\top} &\sim& \pi T \pi^{-1} ([u, v, z, 1]^{\top})\\
    \text{$[\hat{u'}, \hat{v'}, \hat{z'}, 1]^{\top}$} &\sim& \pi T^{-1} \pi^{-1} ([u', v', z', 1]^{\top})
\end{eqnarray*}
where, for instance, $\hat{u}$ denotes the projection of $u$ to the second view,
and $\hat{u'}$ denotes the projection of $u'$ to the first view.
Here, $\pi: \mathbb{R}^4 \rightarrow \mathbb{R}^4$ represents the perspective projection operation
that maps an input homogeneous 3D coordinate $[x, y, z, 1]^{\top}$ in camera coordinates to a pixel position plus depth:
\begin{align}
\pi([x, y, z, 1]^\top) ~=~ \left[\frac{fx}{z}, \frac{fy}{z}, z, 1\right]^{\top} ~=~ [u, v, z, 1]^{\top}
\end{align}
% Given $z$ and $f$, they are related by $[u, v] = [\frac{xf}{z}, \frac{yf}{z}]$.
%%%%
We define a symmetric multi-view consistency loss as:
\begin{align}
L_\text{con} ~=~ \frac{1}{2N}\sum_{i=1}^{N} \Big\| [{u_i},{v_i},{u_i'},{v_i'}]^\top - [\hat{u_i'},\hat{v_i'},\hat{u_i},\hat{v_i}]^\top\Big\|^2
\end{align}
We measure error only in the observable image space $(u, v)$ as opposed to also using $z$, because depth
is never directly observed, and usually has different units compared to $u$ and $v$. Note however that predicting $z$ is critical for us to be able to project points between the two views.
%We choose L2 norm.

Enforcing multi-view consistency is sufficient to infer a consistent set of 2D keypoint positions (and depths) across different views. However, this consistency alone often leads to a degenerate solution where all keypoints collapse to a single location, which is not useful. One can encode an explicit notion of diversity to prevent collapsing, but there still exists infinitely many solutions that satisfy multi-view consistency. Rather, what we need is a notion of optimality for selecting keypoints which has to be defined with respect to some downstream task. For that purpose, we use pose estimation as a task which naturally encourages keypoint separation so as to yield well-posed estimation problems.

%we find that including the downstream objective
%as part of the keypoint selection process naturally encourages keypoint separation.
%Further, there are infinitely many sets of keypoints that  we need a clear %downstream application to decide which keypoints are most useful.

% It is not clear which set of keypoints to select without having a clear downstream application in mind.

% . In our experiments this point often
% lands on the center of mass of the object, which is not useful for most downstream tasks.

\subsection{Relative pose estimation}

One important application of keypoint detection is to recover the relative transformation between a given pair of images. Accordingly, we define a differentiable objective that measures the misfit between the estimated relative rotation $\hat{R}$ (computed via Procrustes' alignment of the two sets of keypoints) and the ground truth $R$. Given the translation equivariance property of our keypoint prediction network (Section \ref{sec:network}) and the view consistency loss above, we omit the translation error in this objective. %\tompson{This is last point is not entirely clear. If they don't understand it the reviewers might think it's because we always center the object in the world coordinates (and we assume t = 0 for all relative camera transforms)... This misconception might lead them to believe that the object must always be centered. While it might be obvious to you, explain (or remind them) how equivariance is ensured when the output is a [N x 3]... Or on futher reading you do describe it in Section 5. so refer them there if you can't summarize it quickly.}
The pose estimation objective is defined as :
\comment{
With the mutli-view consistency loss,
we have some guarantee that the predicted points will be consistent.
However, any point in 3D space that the network can consistently track will satisfy this constraint,
including a point with a fixed length above the object.
Empirically, the network tends to output all keypoints at the center of the object, which satisfy the consistency constraint.
We believe this happens because the center is relatively easy to compute by just taking the average locations of all object's pixels,
and it is the stationary point of rotations, assuming zero-mean center.

The root cause of this problem, rather, is that we lack a metric that defines a ``good'' set of keypoints.

The answer certainly depends on the downstream tasks, which we do not know a priori. However, we propose one kind of metric that 
} % end \comment
\begin{align}
    L_\text{pose} ~=~ 2\arcsin\left(\frac{1}{ 2\sqrt{2}}\left\| {\hat{R}} - {R}\right\|_F \right)
\label{eq:pose}
\end{align}
%\begin{align}
%    L_\text{geo} = \left\| \log {\hat{R}}{R}^\top\right\|^2 
%\end{align}
which measures the angular distance between the optimal least-squares estimate ${\hat{R}}$
computed from the two sets of keypoints, and the ground truth relative rotation matrix ${R}$. % which represents the $3\times 3$ rotation part of the relative transformation $T$ (see \eqref{eq:t}).
Fortunately, we can formulate this objective in terms of fully differentiable operations.

To estimate ${\hat{R}}$, let ${X}$ and ${X'} \in \mathbb{R}^{3 \times N}$ denote two matrices
comprising unprojected 3D keypoint coordinates for the two views. 
In other words, let $X\equiv[X_1, \ldots, X_N]$ and $X_i \equiv (\pi^{-1}p_i)[\text{:}3]$
%\begin{equation}\label{eq:k}
%X_i \equiv (\pi^{-1}p_i)[\text{:}3]
%\end{equation}
, where $[\text{:}3]$ 
% is the de-homogenize function (
returns the first 3 coordinates of its input.
Similarly $X'$ denotes unprojected points in $P'$.
Let $\tilde{X}$ and $\tilde{X}'$ denote the mean-subtracted version of $X$ and $X'$, respectively.
The optimal least-squares rotation ${\hat{R}}$ between the two sets of keypoints is then given by:
\begin{equation}
    {\hat{R}} ~=~ {V} \diag(1, 1, \ldots, \det({V}{U}^\top)) {U}^\top, 
\end{equation}
%where:
%\begin{align*}
%    {U}, {\Sigma}, {V}^\top &= \svd(\tilde{X} \tilde{X}'^{\top})
%\end{align*}
where ${U}, {\Sigma}, {V}^\top = \svd(\tilde{X} \tilde{X}'^{\top})$. This estimation problem to recover $\hat{R}$ is known as the orthogonal Procrustes problem~\cite{schonemann1966procrustes}.
To ensure that $\tilde{X} \tilde{X'}^{\top}$ is invertible and to increase the robustness of the keypoints,
we add Gaussian noise to the 3D coordinates
of the keypoints ($X$ and $X'$) and instead seek the best rotation under some noisy predictions of keypoints. %\tompson{If space persists you might want to formalize this. How did you choose var for the noise? Why does adding noise improve the condition number? (somewhat obvious to me, but maybe not obvious to the reviewer).}
To minimize the angular distance \eqref{eq:pose}, we backpropagate through the SVD operator using matrix calculus~\cite{ionescu2015matrix,giles2008extended}.

Empirically, the pose estimation objective helps significantly in producing a reasonable and natural selection of latent keypoints, leading to the automatic discovery of interesting parts such as the wheels of a car, the cockpit and wings of a plane, or the legs and back of a chair. We believe this is because these parts are geometrically consistent within an object class (e.g., circular wheels appear in all cars), easy to track, and spatially varied, all of which improve the performance of the downstream task.
%\tompson{This needs demystifying a little (comes across a little bit as: "we did this thing, and wow, cool stuff happened"... Cool stuff happens because these landmarks are geometrically consistent within an object class (i.e. all cars have wheels), spatially varied (otherwise inverse above would not exist) and easy to track (high precision) and all these attributes improve the performance of the downstream task (estimating rotation).}

%Note that there are other objectives that can lead to discovery of keypoints. For example, \cite{hinton2011transforming, zhang2018unsupervised} find the optimal set of keypoints that best reconstruct the original input image under a specific decoder network. In this work we are interested in recovering the geometric characteristics of a scene, and accordingly propose the relative pose estimation objective.

% This kind of loss may promote a wide location coverage as such configuration captures more information for reconstruction.
%This reconstruction loss is explored in ~, although under a different context and applications.
%In this work we are interested in recovering the geometric characteristics of a scene, and accordingly we %propose the relative pose estimation objective.

\section{KeypointNet Architecture}\label{sec:network}

% 3D keypoint prediction network

%There are many ways to design a neural network that can output a set of 3D keypoints given an input image.
%For example, one could employ a standard CNN with a fully-connected output layer to regress to an $N\times3$ dimensional output.
%However, 
One important property for the mapping from images to keypoints is translation {\em equivariance} at the pixel level. That is, if we shift the input image, \eg~to the left by one pixel, the output locations of all keypoints should also be changed by one unit.
Training a standard CNN without this property would require a larger training set that contains objects at every possible location, while still providing no equivariance guarantees at inference time.

We propose the following simple modifications to achieve equivariance.
Instead of regressing directly to the coordinate values,
we ask the network to output a probability distribution map $g_i(u, v)$ that represents
how likely keypoint $i$ is to occur at pixel $(u, v)$, with $\sum_{u, v} g_i(u, v)=1$.
%\tompson{Importantly though you use a fully convolutional network which gives translation invariance (up to edge effects). Not all architectures will work and this is probably worth mentioning.}
We use a spatial softmax layer to produce such a distribution over image pixels \cite{goroshin2015learning}.
%% Then, we apply an appropriate weighting to the probability map to recover a pixel coordinate.
We then compute the expected values of these spatial distributions to recover a pixel coordinate:
\begin{equation}
\left[u_i, v_i\right]^{\top} = \sum_{u, v} \left[u \cdot g_i(u, v), v \cdot g_i(u, v)\right]^{\top}\label{eq:kx}
\end{equation}
%%%%
% Suppose the output has a normalized coordinate frame,
% \ie~left bottom pixel is located at (-1, -1) and top right pixel (1, 1).
% Then, let
% \newcommand{\meshgrid}{\mathop{\mathrm{MeshGrid}}}
% \begin{equation}
% W_u, W_v = \meshgrid([-1, 1], [-1, 1]).
% \end{equation}
% Then the $u$ coordinate of $i$th keypoint is computed using an element-wise multiplication,
%by taking an expectation of $W_u$ {\em w.r.t.} the probability distribution $g_i$ and $W_u$,
% and similarly for the $v$ coordinates, \ie~ 
%% \begin{align}\label{eq:kx}
%% u_i &= \sum_{u, v}W_u(u, v) g_i(u, v)\\
%% v_i &= \sum_{u, v}W_v(u, v) g_i(u, v)~.
%% \end{align}
For the $z$ coordinates, we also predict a depth value at every pixel, denoted $d_i(u, v)$, and compute 
\begin{align}\label{eq:kz}
z_i = \sum_{u, v}d_i(u, v) g_i(u, v).
\end{align}
%%%%
To produce a probability map with the same resolution and equivariance property, we use strided-one fully convolutional architectures~\cite{fcn}, also used for semantic segmentation.
%\tompson{Ahhhh... you mention it here. I would move this sentence further up.}
To increase the receptive field of the network, we stack multiple layers of dilated convolutions, similar to~\cite{wavenet}. 

Our emphasis on designing an equivariant network not only helps significantly reduce the number of training examples required to achieve good generalization, but also removes the computational burden of converting between two representations (spatial-encoded in image to value-encoded in coordinates) from the network, so that it can focus on other critical tasks such as inferring depth.

{\bf Architecture details.~} All kernels for all layers are $3\!\times\!3$,
and we stack $13$ layers of dilated convolutions with dilation rates of $1, 1, 2, 4, 8, 16, 1, 2, 4, 8, 16, 1, 1$, all with $64$
output channels except the last layer which has $2N$ output channels, split between $g_i$ and $d_i$.
We use leakyRelu and Batch Normalization~\cite{batchnorm} for all layers except the last layer.
The output layers for $d_i$ have no activation function, and the channels are passed through a spatial softmax to produce $g_i$.
Finally, $g_i$ and $d_i$ are then converted to actual coordinates $p_i$ using Equations \eqref{eq:kx} and \eqref{eq:kz}.

{\bf Breaking symmetry.~} 
Many object classes are symmetric across at least one axis,
\eg~the left side of a sedan looks like the right side flipped.
This presents a challenge to the network because different parts
can appear visually identical, and can only be resolved by understanding global context.
For example, distinguishing the left wheels from the right wheels requires knowing its orientation (\ie~whether it is facing left or right).
Both supervised and unsupervised techniques benefit from some global conditioning to aid in breaking ties
and to make the keypoint prediction more deterministic.

% this is not as difficult to deal with because each keypoint is semantically defined beforehand (\eg~front-left wheel) and there is a clear supervision that moves the prediction toward its correct position. For unsupervised techniques, however, a keypoint can be directed towards two different, but visually similar parts across the symmetry plane and get stuck in a local minimum.

To help break symmetries,
one can condition the keypoint prediction on some coarse quantization of the pose.
Such a coarse-to-fine approach to keypoint detection is discussed in more depth in~\cite{tulsiani2015viewpoints}.
% the keypoint prediction on the object orientation,
% which signals the keypoint to select only one of the two gradient directions.
One simple such conditioning is a binary flag that indicates whether
the dominant direction of an object is facing left or right.
This dominant direction comes from the ShapeNet dataset we use
(\secref{sec:trainingdata}), where the 3D models are consistently oriented.
To infer keypoints without this flag at inference time,
we train a network with the same architecture, although half the size,
to predict this binary flag.

In particular, we train this network to predict the projected pixel locations of
two 3D points $[1, 0, 0]$ and $[-1, 0, 0]$, transformed into each view in a training pair.
These points correspond to the front and back of a normalized object.
This network has a single $L_2$ loss between the predicted and the ground-truth locations.
The binary flag is 1 if the $x-$coordinate of the projected pixel of the first point is greater than that of the second point.
This flag is then fed into the keypoint prediction network. %\mohammad{did not understand this part: concatenated to 4$^{th}$ channel of the input image.}

%One advantage of predicting a binary flag this way over using a standard CNN with a sigmoid output is the translation invariance. 

% There are many advantages of predicting a binary flag this way over using a standard CNN with a sigmoid output or cross-entropy output. One is that we achieve translation invariance due to the equivariant predictions of the two points.

\section{Additional Keypoint Characteristics}

In addition to the main objectives introduced above, there are common, desirable characteristics of keypoints that can benefit many possible downstream tasks, in particular:
\begin{packed_item}
\setlength\topsep{0em}
\item No two keypoints should share the same 3D location.
\item Keypoints should lie within the object's silhouette.
\end{packed_item}

{\bf Separation loss} penalizes two keypoints if
they are closer than a hyperparameter $\delta$ in 3D:
\begin{align}
\begin{split}
    L_\text{sep} ~=~ \frac{1}{N^2}\sum_{i=1}^N \sum_{j\neq i}^N \max\left(0, \delta^2 - \left\|X_i - X_j\right\|^2\right)
\end{split}
\end{align}
% where $k_i$ denotes the 3D coordinate of the $i$th keypoint in the first view~\eqref{eq:k}.
Unlike the consistency loss, this loss is computed in 3D to allow multiple keypoints to occupy the same pixel location as long as they have different depths. We prefer a robust, bounded support loss over an unbounded one (\eg~exponential discounting) because it does not exhibit a bias towards certain structures, such as a honeycomb, or towards placing points infinitely far apart. Instead, it encourages the points to be sufficiently far from one another.

Ideally, a well-distributed set of keypoints will
automatically emerge without constraining the distance of keypoints. However, in the absence of keypoint location supervision, our objective with latent keypoints can converge to a local minimum with two keypoints collapsing to one. The main goal of this separation loss is to prevent such degenerate cases, and not to directly promote separation. 

\textbf{Silhouette consistency} encourages the keypoints to lie within the silhouette of the object of interest. As described above, our network predicts $(u_i, v_i)$ coordinates of the $i^\text{th}$ keypoint via
a spatial distribution, denoted $g_i(u, v)$, over possible keypoint positions.
One way to ensure silhouette consistency, is by \emph{only} allowing a non-zero probability inside the silhouette of the object, as well as encouraging the spatial distribution to be concentrated,
\ie~uni-modal with a low variance.

During training, we have access to the binary segmentation mask of the object $b(u, v) \in \{0, 1\}$ in each image, where $1$ means foreground object. The silhouette consistency loss is defined as
\begin{align}
L_\text{obj} ~=~ \frac{1}{N}\sum_{i=1}^N -\log \sum_{u, v}b(u, v) g_i(u, v)
\end{align}
Note that this binary mask is only used to compute the loss and not used at inference time. This objective incurs a zero cost if all of the probability mass lies within the silhouette. We also include a term to minimize the variance of each of the distribution maps:
\begin{align}
L_\text{var} = \frac{1}{N} \sum_{i=1}^N \sum_{u, v} g_i(u, v) \Big\| [u, v]^\top - [u_i, v_i]^\top \Big\|^2
\end{align}
This term encourages the distributions to be peaky, which has the added benefit of helping keep their means within the silhouette in the case of non-convex object boundaries.
%\tompson{These segmentation masks are more labels that you're using and that weren't mentioned in the introduction... It also makes baseline comparison less fair (AFAIK you don't inlclude this loss when training the supervised network, correct me if I'm wrong). A careful reviewer might complain that you didn't mention this.}

\section{Experiments} \label{experiments}
%This section describes our evaluations and results on various object classes.
\newcommand\Tstrut{\rule{0pt}{2.6ex}}         % = `top' strut
\newcommand\Bstrut{\rule[-0.9ex]{0pt}{0pt}}   % = `bottom' strut

\paragraph{Training data.}\label{sec:trainingdata}
Our training data is generated from ShapeNet \cite{chang2015shapenet}, a large-scale database of approximately 51K 3D models across 270 categories. We create separate training datasets for various object categories, including car, % (ShapeNet class ID:02958343), 
chair,
% (03001627,02738535),
and plane.
% (02691156). 
For each model in each category, we normalize the object so that the longest dimension lies in $[-1, 1]$, and render 200 images of size $128\times 128$ under different viewpoints to form 100 training pairs. The camera viewpoints are randomly sampled around the object from a fixed distance, all above the ground with zero roll angle. We then add small random shifts to the camera positions.
%We render each object with one hemisphere light directly above the object using Blender. 

\setlength{\belowcaptionskip}{-10pt}
\begin{figure*}[t] \center
	\includegraphics[width=1.0\textwidth]{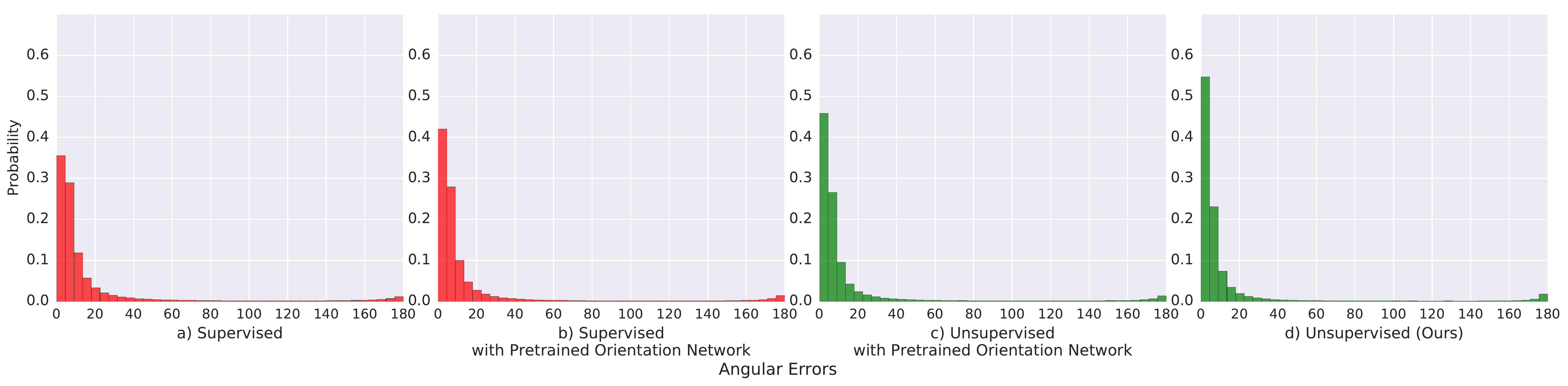}
	\vskip -.1em
	\caption{Histogram plots of angular distance errors, average across car, plane, and chair categories, between the ground-truth relative rotations and the least-squares estimates computed from two sets of keypoints predicted from test pairs. a) is a supervised method trained with a single $L_2$ loss between the pixel location prediction to the human labels. b) is the same as a) except the network is given an additional orientation flag predicted from a pretrained orientation network. c) is our network that uses the same pretrained orientation network as b), and d) is our unsupervised method trained jointly (the orientation and keypoint networks).\label{fig:supervised_comparison} 
	%\tompson{nit: It's probably too late, but it might be better to combine columns 1 and 2 and combine columns 3 and 4 so there are 2 figures (there are lots of visually pleasing ways of drawing bar graphs on a single axis). Otherwise the text on these figures is really, really tiny.}
	% Supasorn: Good and interesting idea. Too late now. Have 3 more histograms in supp.
	}
\end{figure*}

{\bf Implementation details.~}
We implemented our network in TensorFlow~\cite{tensorflow2015-whitepaper}, and trained with the Adam optimizer
with a learning rate of $10^{-3}, \beta_1=0.9 ,\beta_2=0.999$, and a total batch size of $256$.
%For all categories,
We use the following weights for the losses:
$(\alpha_\text{con},\alpha_\text{pose},\alpha_\text{sep},\alpha_\text{obj}) = (1, 0.2, 1.0, 1.0)$.
We train the network for $200K$ steps using synchronous training with $32$ replicas.

% \alpha_\text{sym} = 1.0

\subsection{Comparison with a supervised approach}

To evaluate against a supervised approach, we collected human landmark labels for three object categories (cars, chairs, and planes) from ShapeNet using Amazon Mechanical Turk. For each object, we ask three different users to click on points corresponding to reference points shown as an example to the user. These reference points are based on the Pascal3D+ dataset (12 points for cars, 10 for chairs, 8 for planes). We render the object from multiple views so that each specified point is facing outward from the screen. We then compute the average pixel location over user annotations for each keypoint, and triangulate corresponding points across views to obtain 3D keypoint coordinates.

For each category, we train a network with the same architecture as in Section~\ref{sec:network} using the supervised labels to output keypoint locations in normalized coordinates $[-1, 1]$, as well as depths, using an $L_2$ loss to the human labels. We then compute the angular distance error on 10\% of the models for each category held out as a test set. (This test set corresponds to 720 models of cars, 200 chairs, and 400 planes. Each individual model produces 100 test image pairs.) In Figure \ref{fig:supervised_comparison}, we plot the histograms of angular errors of our method vs.\ the supervised technique trained to predict the same number of keypoints, and show error statistics in Table \ref{tab:rotation_error}. For a fair comparison against the supervised technique, we provide an additional orientation flag to the supervised network. This is done by training another version of the supervised network that receives the orientation flag predicted from a pre-trained orientation network. Additionally, we tested a more comparable version of our unsupervised network where we use and fix the same pre-trained orientation network during training. The mean and median accuracy of the predicted orientation flags on the test sets are as follows: cars: ($96.0\%$, $99.0\%$), planes: ($95.5\%$, $99.0\%$), chairs: ($97.1\%$, $99.0\%$). 

Our unsupervised technique produces lower mean and median rotation errors than both versions of the supervised technique. Note that our technique sometimes incorrectly predicts keypoints that are \ang{180} from the correct orientation due to incorrect orientation prediction.

\definecolor{light_gray}{rgb}{0.8,0.8,0.8}
\newcommand{\tspace}{\hspace{0.7em}}

\begin{table}[t]
\small
\centering 
%\vspace*{-.3em}
\setlength{\tabcolsep}{0.26em}
\begin{tabular}{lccccccccc} 
%& \multicolumn{4}{c}{RMSE}\\
\hline
\Xhline{3\arrayrulewidth}
 & \multicolumn{3}{c}{Cars} & \multicolumn{3}{c}{Planes} & \multicolumn{3}{c}{Chairs} \Tstrut \\
Method & \footnotesize{Mean} & \footnotesize{Median} & \footnotesize{3D-SE} \tspace & \tspace \footnotesize{Mean} & \footnotesize{Median} & \footnotesize{3D-SE} \tspace  & \tspace \footnotesize{Mean} & \footnotesize{Median} & \footnotesize{3D-SE}  \Bstrut\\
\Xhline{2\arrayrulewidth}
a) Supervised & 16.268 & 5.583 & 0.240  & \tspace 18.350 & 7.168 & 0.233 & \tspace 21.882 & 8.771 & 0.269 \Tstrut\\
\arrayrulecolor{light_gray}\hline
b) Supervised with & 13.961 & 4.475 & 0.197 & \tspace 17.800 & 6.802 & 0.230 & \tspace 20.502 & 8.261 & 0.248 {\rule{0pt}{2ex}} \\ \footnotesize{\ \ \ \ pretrained O-Net} & & & & & & & & \Bstrut\\
\arrayrulecolor{black}\hline \Tstrut
c) Ours with \\ \footnotesize{\ \ \ \ pretrained O-Net} & 13.500 & 4.418 & 0.165 & \tspace 18.561 & 6.407 & 0.223 & \tspace 14.238 & 5.607 & 0.203  \\
\arrayrulecolor{light_gray}\hline
d) \textbf{Ours} & 11.310 & 3.372 & 0.171 & \tspace 17.330 & 5.721 & 0.230 & \tspace 14.572 & 5.420 & 0.196{\rule{0pt}{2ex}}\Bstrut\\
\arrayrulecolor{black}
\Xhline{2\arrayrulewidth}
\vspace*{-.3em}
\end{tabular} \caption{Mean and median angular distance errors between the ground-truth 
rotation and the Procrustes estimate computed from two sets of predicted keypoints on test pairs. O-Net is the network that predicts a binary orientation. 3D-SE is the standard errors described in Section \ref{sec:keypoint_consistency}.
% \mohammad{I don't see the point of the supervised line. Why not just compare against supervised + O-net? Are we trying to argue that O-net is important? But that indicates something is wrong with supervised.} 
% O-net is an idea introduced in this paper. A proper, most basic baseline is supervised without orientation prediction. Without it, ppl might also ask if adding the flag actually hurts supervised methods. 
} %title of the table
\label{tab:rotation_error}
\vspace*{-1.5em}
\end{table}

\paragraph{Keypoint location consistency.~}\label{sec:keypoint_consistency}
To evaluate the consistency of predicted keypoints across views, we transform the keypoints predicted for the same object under different views to object space using the known camera matrices used for rendering. Then we compute the standard error of 3D locations for all keypoints across all test cars (3D-SE in Table \ref{tab:rotation_error}). To disregard outliers when the network incorrectly infers the orientation, we compute this metric only for keypoints whose error in rotation estimate is less than \ang{90} 
% as an approximation 
(left halves of the histograms in Figure \ref{fig:supervised_comparison}), for both the supervised method and our unsupervised approach.

%\begin{figure}[t] \center
%	\vskip -.6em
	
	%\vskip -.6em
%\end{figure}
\begin{figure}[t] \center
    %\vskip 0em
    \includegraphics[width=0.98\textwidth]{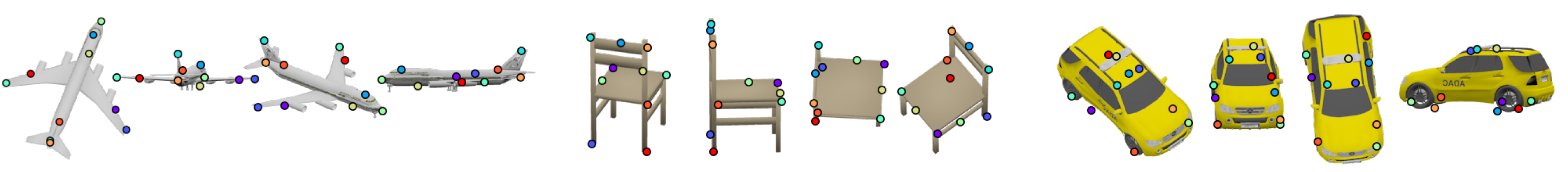}
    %\vskip -.6em
	\caption{Keypoint results on single objects from different views. Note that these keypoints are  predicted consistently across views even when they are completely occluded. (\eg~the red point that tracks the back right leg of the chair.) \label{fig:cpc} %\tompson{I think you should put a hlink to the URL again in this caption.}
	Please see \link for visualizations.
	}
	
	\vskip 3em
    \includegraphics[width=0.98\textwidth]{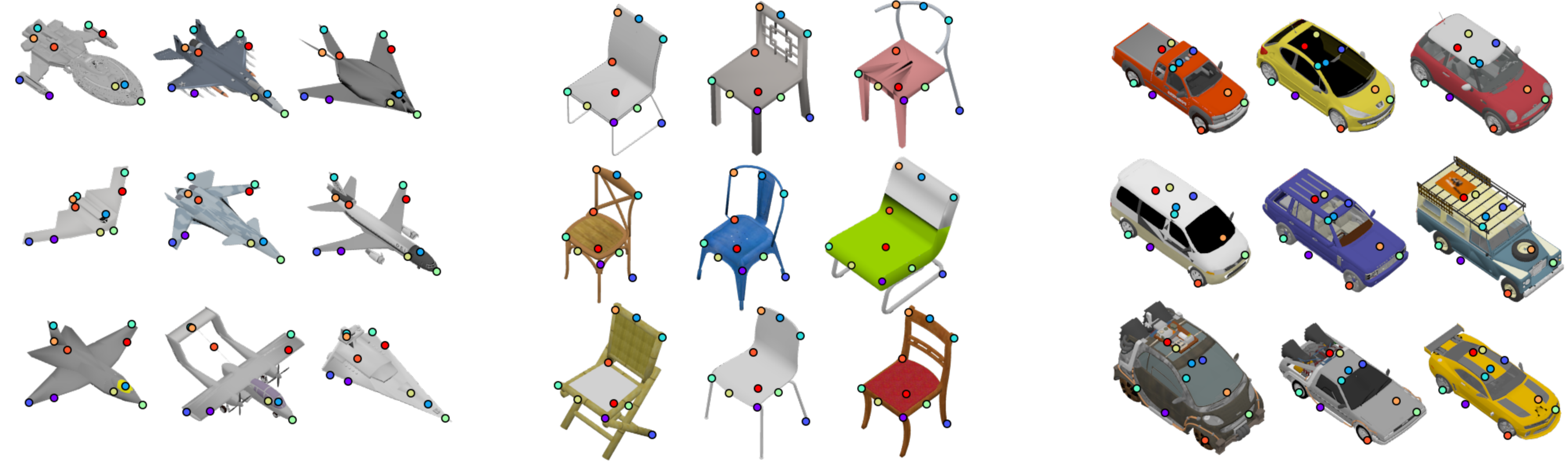}
    \vskip -.1em
    \caption{Results on ShapeNet~\cite{chang2015shapenet} test sets for cars, planes, and chairs. Our network is able to generalize across unseen appearances and shape variations, and consistently predict occluded parts such as wheels and chair legs.
    %\mohammad{we may want to make it clear that colored keypoints across different categories don't correspond to each other.}
    \label{fig:cpcs}}
    \vskip -.1em
\end{figure}

% \subsubsection{depth prediction accuracy}

%
%\subsection{Comparison to correspondence finding algorithms}
%Point out the distinction between correspondence finding algorithms which need 2 %images and correspondence is defined with respect to the given pair.
%\begin{figure} \center
	%\includegraphics[width=0.48\textwidth]{figures/compare_corresalgo.pdf}
	%\caption{Comparison against other correspondence finding algorithms. %\label{fig:overview}}
%\end{figure}

\vspace*{-.2cm}
\subsection{Generalization across views and instances}
\vspace*{-.1cm}
In this section, we show qualitative results of our keypoint predictions on test cars, chairs, and planes using a default number of 10 keypoints for all categories. (We show results with varying numbers of keypoints in the Appendix.) In Figure \ref{fig:cpc}, we show keypoint prediction results on single objects from different views. Some of these views are quite challenging such as the top-down view of the chair. However, our network is able to infer the orientation and predict occluded parts such as the chair legs. In Figure \ref{fig:cpcs}, we run our network on many instances of test objects. Note that during training, the network only sees a pair of images of the same model, but it is able to utilize the same keypoints for semantically similar parts across all instances from the same class. For example, the blue keypoints always track the cockpit of the planes. In contrast to prior work \cite{thewlis2017unsupervised, hinton2011transforming, zhang2018unsupervised} that learns latent representations by training with restricted classes of transformations, such as affine or 2D optical flow, and demonstrates results on images with small pose variations, we learn through physical 3D transformation and are able to produce a consistent set of 3D keypoints from any angle. Our method can also be used to establish correspondence between two views under out-of-plane or even 180$^\circ$ rotations when there is no visual overlap.

\begin{minipage}{0.6\textwidth}
{\bf Failure cases.~}
When our orientation network fails to predict the correct orientation, the output keypoints will be flipped as shown in Figure \ref{fig:fail}. This happens for cars whose front and back look very similar, or for unusual wing shapes that make inference of the dominant direction difficult.
\end{minipage}%
\hfill
\begin{minipage}{0.39\textwidth}
    \includegraphics[width=1\textwidth]{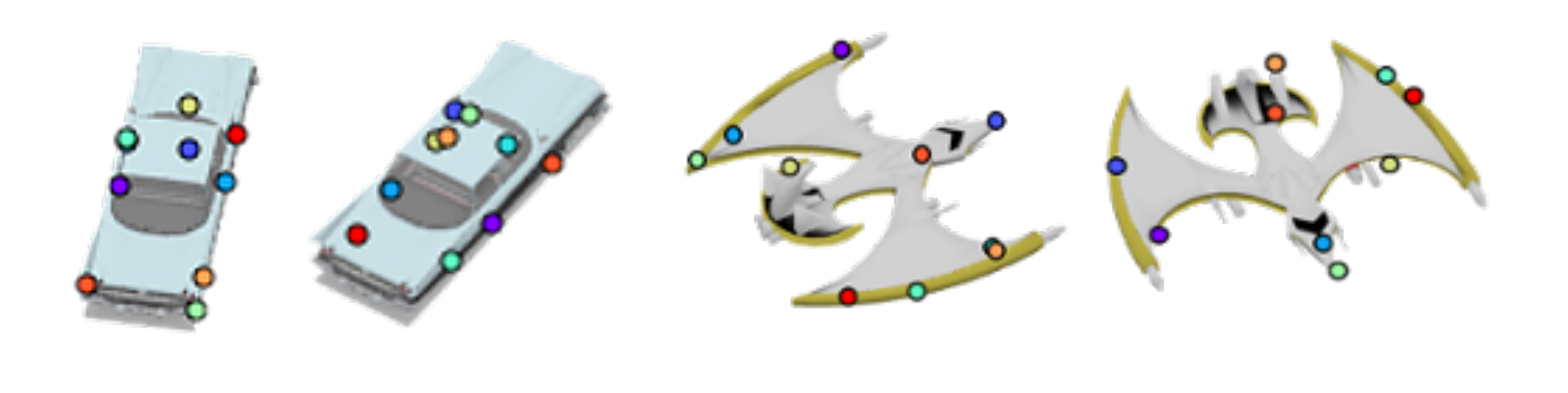}
    \captionof{figure}{Failure cases. \label{fig:fail}}
    \vspace*{-.2cm}
\end{minipage}%

%\textbf{Result on real images.~} We show a proof-of-concept inference result on real images from the Pascal3D+~\cite{xiang2014beyondpascal} dataset (Figure \ref{fig:fail}), which suggests that our proposed technique can be applicable across domains.

%\subsection{Results on Real Images} MOVE TO DISCUSSION

\vspace*{-.1cm}
\section{Discussion \& Future work} \label{conclusion}

%We examine end-to-end method for unsupervised discovery of 3D landmarks from monocular RGB images.

We explore the possibility of optimizing a representation based on a {\em sparse}
set of keypoints or landmarks, without access to keypoint annotations, but rather based on
an end-to-end geometric reasoning framework. We show that,
indeed, one can discover consistent keypoints across multiple views and object instances by
adopting two novel objective functions: a relative pose estimation loss and a multi-view consistency objective.
%\tompson{"keypointness"? A little to vague. Why not be more specific? geometric consistency losses?}
Our translation equivariant architecture is able to generalize to unseen object
instances of ShapeNet categories~\cite{chang2015shapenet}.
Importantly, our discovered keypoints outperform those from a direct supervised learning
baseline on the problem of rigid 3D pose estimation.

%~\cite{johnson2017driving}, domain randomization \cite{tremblay2018training, tobin2017domain}, or GANs \cite{bousmalis2017unsupervised, tzeng2017adversarial}.
We present preliminary results on the transfer of the learned keypoint detectors to real world images by training on
ShapeNet images with random backgrounds (see supplemental material). Further improvements may be achieved by leveraging recent work in domain adaptation~\cite{johnson2017driving,tremblay2018training, tobin2017domain,bousmalis2017unsupervised, tzeng2017adversarial}.
% Applying this technique to real-world images is an exciting next step. One possible way to accomplish this is to apply domain adaptation techniques that allow a network trained entirely on synthetic data to perform inference on real images. We show a proof-of-concept result on real images by training our network with random backgrounds (see appendix). Further improvements can be achieved by following recent successes in this field, such as using photorealistic rendering \cite{johnson2017driving}, domain randomization \cite{tremblay2018training, tobin2017domain}, or GANs \cite{bousmalis2017unsupervised, tzeng2017adversarial}.
Alternatively, one can train KeypointNet directly on real images provided relative pose labels.
Such labels may be estimated automatically using Structure-from-Motion \cite{longuet1981computer}.
Another interesting direction would be to jointly solve for the relative transformation or rely on a coarse pose initialization, inspired by \cite{triggs1999bundle}, to extend this framework to objects that lack 3D models or pose annotations. 

Our framework could also be extended to handle an arbitrary number of keypoints. For example, one could predict a confidence value for each keypoint, then threshold to identify distinct ones, while using a loss that operates on unordered sets of keypoints. Visual descriptors could also be incorporated under our framework, either through a post-processing task or via joint end-to-end optimization of both the detector and the descriptor.

\section{Acknowledgement} We would like to thank Chi Zeng who helped setup the Mechanical Turk tasks for our evaluations.

\comment{
Our network can be applied to automatically establish ShapeNet's 3D correspondence among objects within each shape category, which never exists before in the dataset, by ray casting 3D keypoints back to the 3D mesh.
%or finding the nearest vertices.
%\tompson{2 points. 1. "In it's current setup"... it's the only setup. consider removing. 2. Last part is somewhat misleading. We backprop to renderings of the 3D meshes. Not the geometry itself.}
% Supasorn: 1. Done, 2. Our keypoint prediction is exactly what mturkers do. So we can ray cast back to get 3d vertex. Another way is just to find nearest vertex because we have 3D keypoints.
Such keypoints are very useful for graphics applications, such as 3D object retrieval, or can be used as initialization for dense correspondence matching. **
}
%We believe our proposed framework is applicable to many downstream vision tasks.

%\tompson{it feels a little odd finishing on a promise to do more work.  IMO you should move the 3rd paragraph to the end ("Our network can be applied to automatically..."), and finish on a stronger note.}

%Extending KeypointNet to non-rigid human pose estimation problem, or SLAM to discover reliable keypoints to track are also promising. 

\bibliography{paper}
\bibliographystyle{plain}

\ifdefined\ArXiv 
\newpage 
\appendix \section{Histograms for individual categories}
We show histograms similar to Figure 2 in the paper for individual object categories.

\setlength{\belowcaptionskip}{-10pt}
\begin{figure}[htp] \center
	\includegraphics[width=0.99\textwidth]{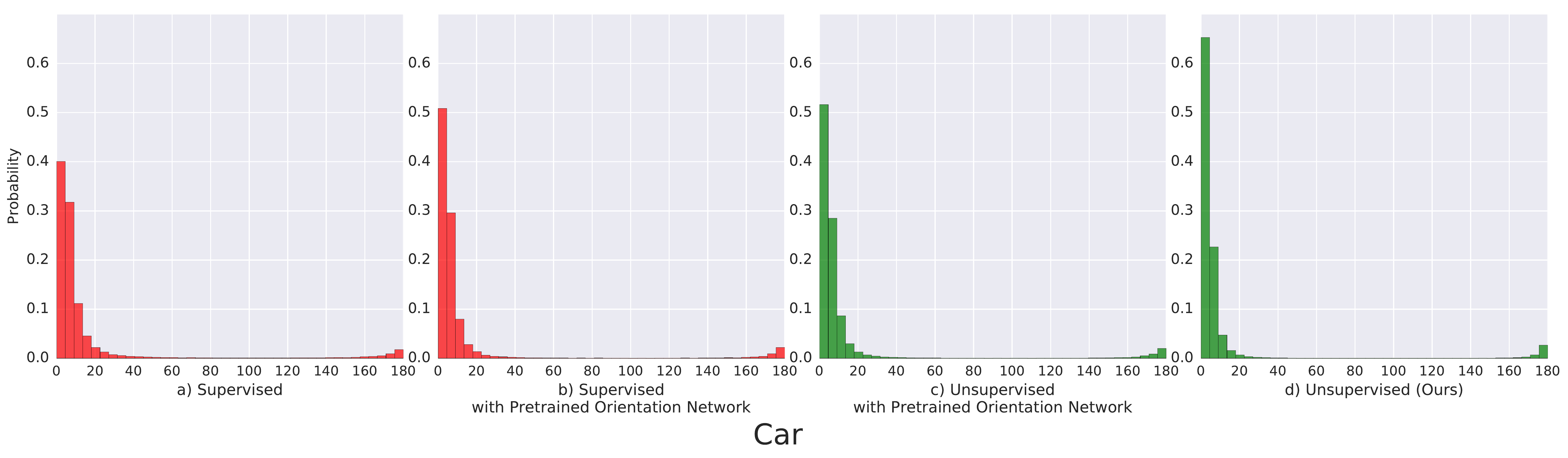}
	\includegraphics[width=0.99\textwidth]{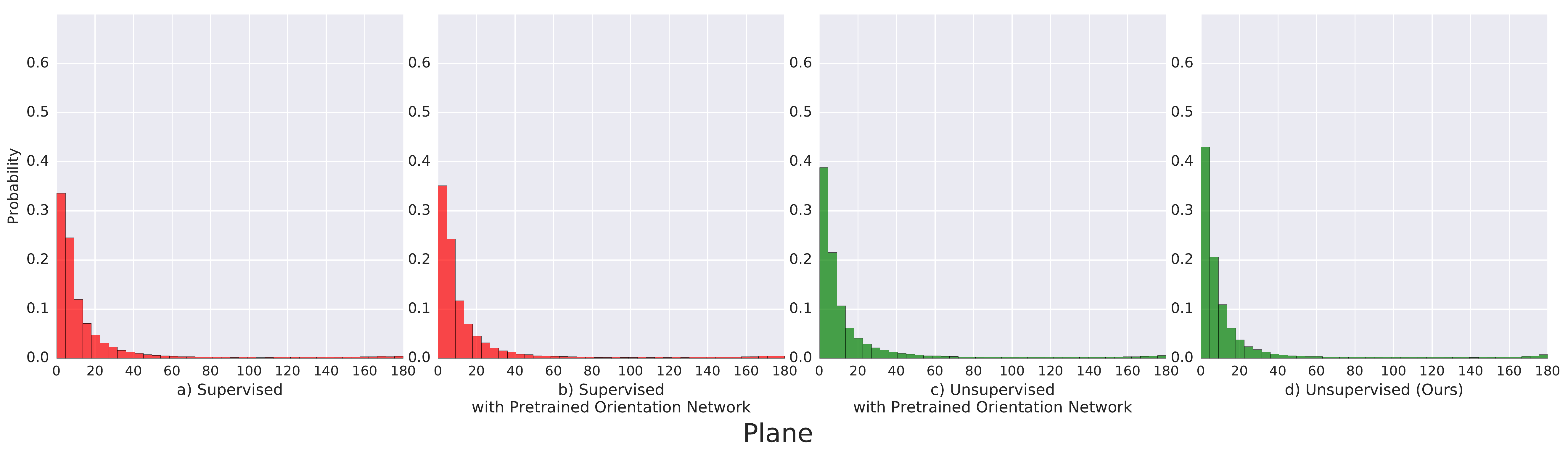}
	\includegraphics[width=0.99\textwidth]{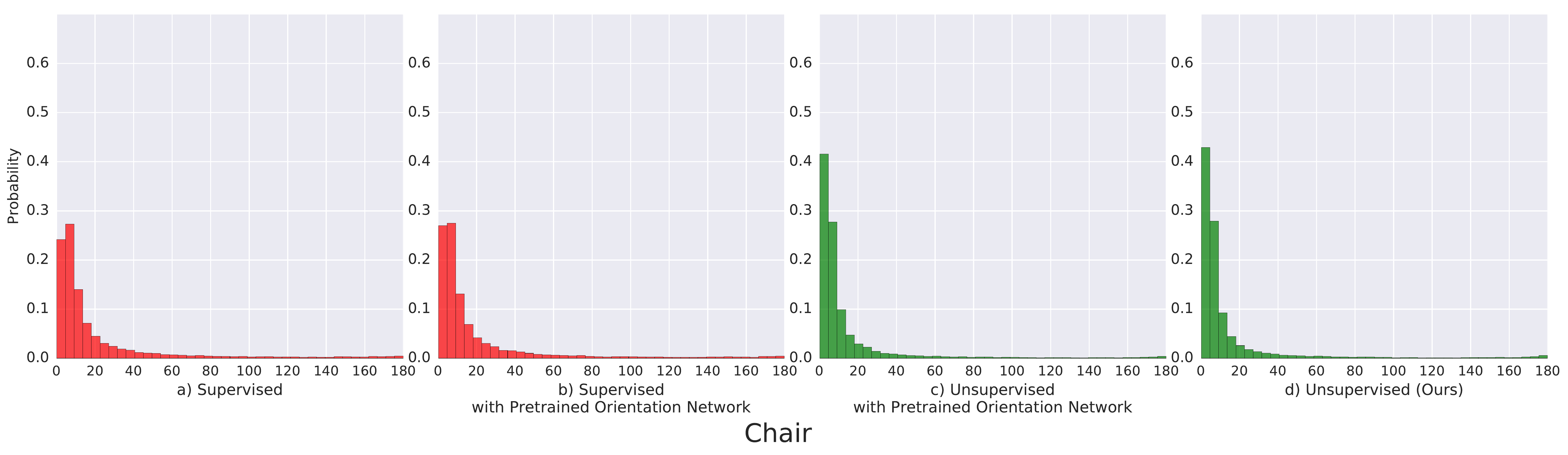}
	\vskip -.5em
	\caption{Histogram plots of angular distance errors between the ground-truth relative rotations and the least-squares estimates computed from two sets of keypoints predicted from test pairs. a) is a supervised method trained with a single $L_2$ loss between the pixel location prediction to the human labels. b) is the same as a) except the network is given an additional orientation flag predicted from a pretrained orientation network. c) is our network that uses the same pretrained orientation network as b), and d) is our unsupervised method trained jointly (the orientation and keypoint networks).\label{fig:supervised_comparison_hist}}
\end{figure}

\section{Ablation study}
We present an ablation study for the primary losses as well as how their weights affect the results (Figure \ref{fig:ablation}).

\begin{figure}[htp]\center
    \includegraphics[width=0.7\textwidth]{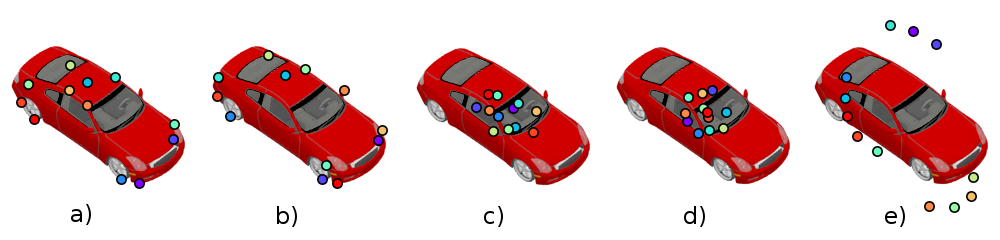}
	\caption{An ablation study for the losses. a) Our baseline model. b) and c) use twice the noise (0.2)
	and no noise respectively in the pose estimation loss. d) removes the pose estimation loss.
	e) removes the silhouette loss. \label{fig:ablation}}
\end{figure}

\textbf{Removing multi-view consistency loss.~} This causes some of the keypoints to move around when the viewing angle changes, and not track onto any particular part of the object. The pose estimation loss alone may only provide a strong gradient for a number of keypoints as long as they give a good rotation estimate, but it does not explicitly force every point to be consistent.

\textbf{Pose estimation loss \& Noise.~} Removing pose estimation loss completely leads the network to place keypoints near the center of an object, which is the area with the least rotation motion, and thus least pixel displacement under different views. Increasing the noise that is added to the keypoints for rotation estimation encourages the keypoints to be spread apart from the center.

\textbf{Removing silhouette consistency.~} This causes the keypoints to lie outside the object. Interestingly, the keypoints still satisfy multi-view consistency, and lie on a virtual 3D space that rotates with the object.

\section{Results on deformed object}
To evaluate the robustness of these keypoints under shape variations such as the length of the car, and whether the network uses local features to detect local parts as opposed to placing keypoints on a regular rigid structure, we run our network on a non-rigidly deformed car in Figure \ref{fig:deformed}. Here we show that the network is able to predict where the wheels are and the overall deformation of the car structure.

\begin{figure}[htp]\center

	\includegraphics[width=0.7\textwidth]{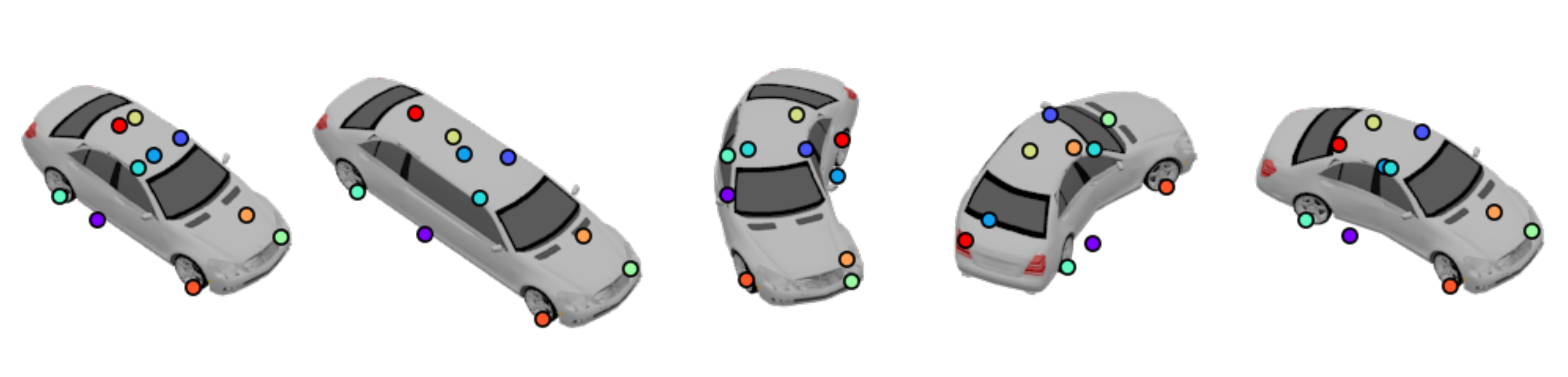}
	\caption{Results on a non-rigidly deformed car. \label{fig:deformed}}
\end{figure}

\section{Results using different numbers of keypoints}
We trained our network with varying number of keypoints $\{3, 5, 8, 10, 15, 20\}$. The network starts by discovering the most prominent components such as the head and wings, then gradually tracks more parts as the number increases. 

\begin{figure}[htp]\center
	\includegraphics[width=0.95\textwidth]{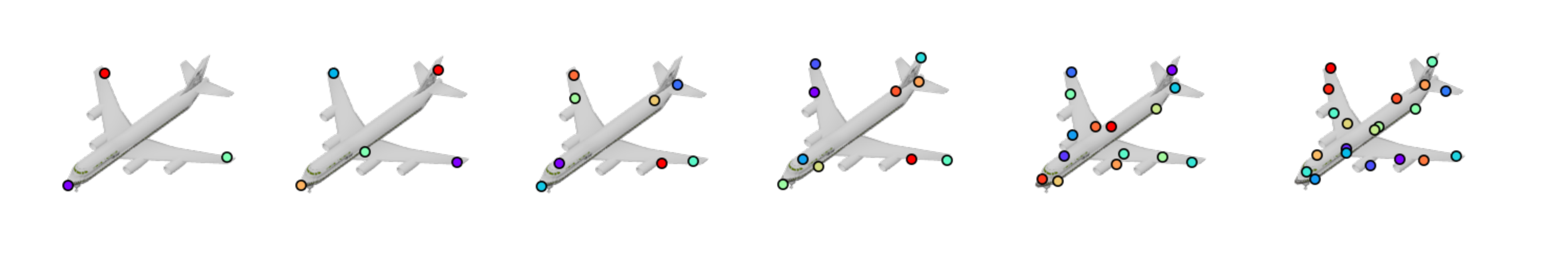}
	\caption{Results using networks trained to predict different numbers of keypoints. (Colors do not correspond across results as they are learned independently.)\label{fig:differentpoints}}
\end{figure}

\section{Proof-of-concept results on real-world images}
To predict keypoints on real images, we train our network by adding random backgrounds, taken from SUN397 dataset \cite{xiao2010sun}, to our rendered training examples. Surprisingly, such a simple modification allows the network to predict keypoints on some cars in ImageNet. We show a few hand-picked results as well as some failure cases in Figure \ref{fig:real}. The network especially has difficulties dealing with large perspective distortion and cars that have strong patterns or specular highlights. 

\begin{figure}[htp]\center
	\includegraphics[width=0.98\textwidth]{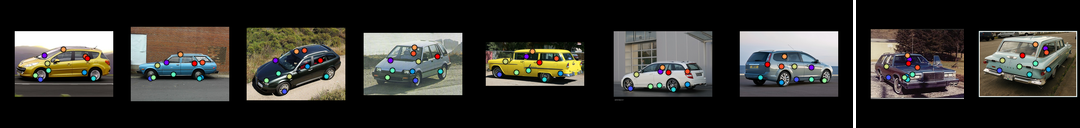}
	\caption{Proof-of-concept results on real images.\label{fig:real}}
\end{figure}

\else 
\fi

\end{document}